\documentclass[runningheads]{llncs}

\usepackage{graphicx}
\usepackage{comment}
\usepackage{amsmath,amssymb} % define this before the line numbering.
\usepackage{color}
\usepackage{cite}

\usepackage{wrapfig}

\usepackage{epsfig}
\usepackage{booktabs}
\usepackage{multirow}
\usepackage[export]{adjustbox}
\usepackage{float}
\usepackage{dashrule}
\usepackage{arydshln}
\usepackage[utf8x]{inputenc}
\usepackage{subcaption}
\usepackage{algorithm}
\usepackage{algorithmic}

\usepackage{xcolor} 
\renewcommand\fbox{\fcolorbox{gray}{white}}
\newcommand{\etal}{et al.}

\begin{document}
% \linenumbers
\pagestyle{headings}
\mainmatter
\def\ECCVSubNumber{5063}  % Insert your submission number here

%%% TITLE
%HoughNet: Integrating near and long-range evidence for visual recognition VEYA \\ 

\title{HoughNet: Integrating near and long-range evidence for bottom-up object detection} % Replace with your title

\begin{comment}
\titlerunning{ECCV-20 submission ID \ECCVSubNumber}
\authorrunning{ECCV-20 submission ID \ECCVSubNumber}
\author{Anonymous ECCV submission}
\institute{Paper ID \ECCVSubNumber}
\end{comment}
%******************

%\begin{comment}
\titlerunning{HoughNet}
\author{Nermin Samet\inst{1} \and
Samet Hicsonmez\inst{2} \and
Emre Akbas\inst{1}}
\authorrunning{N. Samet et al.}
\institute{Department of Computer Engineering, Middle East Technical University 
\email{\{nermin,emre\}@ceng.metu.edu.tr}   \and
Department of Computer Engineering, Hacettepe University
\email{\{samethicsonmez\}@hacettepe.edu.tr}  }

%\end{comment}
%******************
\maketitle

\begin{abstract}

This paper presents HoughNet, a one-stage, anchor-free, voting-based, bottom-up object detection method. Inspired by the Generalized Hough Transform, HoughNet determines the presence of an object at a certain location by the sum of the votes cast on that location. Votes are collected from both near and long-distance locations based on a log-polar vote field. Thanks to this voting mechanism, HoughNet is able to integrate both near and long-range, class-conditional evidence for visual recognition, thereby generalizing and enhancing current object detection methodology, which typically relies on only local evidence.  On the COCO dataset, HoughNet’s best model achieves $46.4$ $AP$ (and $65.1$ $AP_{50}$), performing on par with the state-of-the-art in bottom-up object detection and outperforming most  major one-stage and two-stage methods. We further validate the effectiveness of our proposal in another task, namely, ``labels to photo’’ image generation by integrating the voting module of HoughNet to two different GAN models and showing that the accuracy is significantly improved in both cases. Code is available at \url{https://github.com/nerminsamet/houghnet}. 
\end{abstract}

%%%%%%%%% BODY TEXT
\section{Introduction}
Deep learning has brought on remarkable improvements in object detection. Performance on widely used benchmark datasets, as measured by mean average-precision (mAP), has at least doubled (from 0.33 mAP \cite{voc-release5} \cite{felzenszwalb2009object} to 0.80 mAP on PASCAL VOC~\cite{resnet}; and from 0.2 mAP~\cite{mscoco} to around 0.5 mAP on COCO~\cite{retinanet}) in comparison to the previous generation (pre-deep-learning, shallow) methods. Current state-of-the-art, deep learning based object detectors \cite{faster, ssd, yolo2, retinanet} predominantly follow a top-down approach where objects are detected holistically via rectangular region classification. This was not the case with the pre-deep-learning methods. The bottom-up approach was a major research focus as exemplified by the prominent voting-based (the Implicit Shape Model \cite{ism}) and part-based (the Deformable Parts Model \cite{dpm-pami}) methods. However, today, among deep learning based object detectors, the bottom-up approach has not been sufficiently explored with a few exceptions (e.g. CornerNet \cite{cornernet}, ExtremeNet \cite{extremenet}). %We argue that the bottom-up approach offers promising ways to improve object detection. 

%%%%%%%%%%% Figure Teaser %%%%%%%%

\setlength{\fboxsep}{.15pt}% border’la image arasindaki uzaklik
\setlength{\fboxrule}{.5pt}% border’in kalinligi

\setlength\intextsep{0pt}
\begin{wrapfigure}[18]{r}{0.5\textwidth}

\captionsetup[subfigure]{labelformat=empty}
\centering
\begin{subfigure}[b]{0.21\textwidth} \fbox{\includegraphics[width=1.0\textwidth]{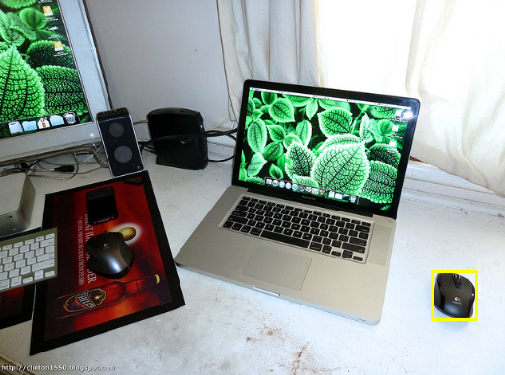}}
\end{subfigure}
\begin{subfigure}[b]{0.21\textwidth} \fbox{\includegraphics[width=1.0\textwidth]{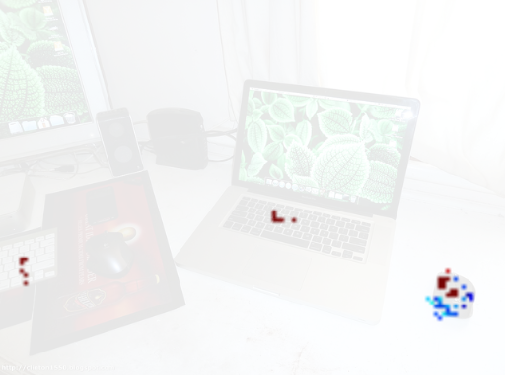}}
\end{subfigure}
\begin{subfigure}[b]{0.028\textheight} \includegraphics[width=1.0\textwidth]{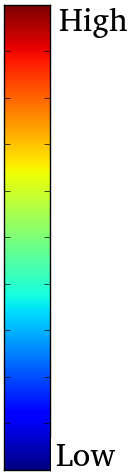}
\end{subfigure}
\caption{(Left) A sample ``mouse’’ detection, shown with yellow bounding box, by HoughNet. (Right) The locations that vote for this detection.  Colors indicate vote strength. In addition to the local votes originating from the mouse itself, there are strong votes from nearby ``keyboard’’ objects, which shows that HoughNet is able to utilize both short and long-range evidence for detection. More examples can be seen in Fig.~\ref{fig:vis-results}}
\label{fig:teaser}
\end{wrapfigure}

%Yaklasimin ve yontemin anlatilmasi
In this paper, we propose HoughNet, a one-stage, anchor-free, voting-based, bottom-up object detection method. HoughNet is based on the idea of voting, inspired by the Generalized Hough Transform~\cite{hough1959,ght1981}. In its most generic form, the goal of GHT is to detect a whole shape based on its parts. Each part produces a hypothesis, i.e. casts its vote, regarding the location of the whole shape. Then, the location with the most votes is selected as the result. Similarly, in HoughNet,  the presence of an object belonging to a certain class at a particular location is determined by the sum of the class-conditional votes cast on that location (Fig.~\ref{fig:teaser}). HoughNet processes the input image using a convolutional neural network to produce an intermediate score map per class. Scores in these maps indicate the presence of visual structures that would support the detection of an object instance. These structures could be object parts, partial objects or patterns belonging to the same or other classes. We name these  score maps as ``visual evidence’’ maps. Each spatial location in a visual evidence map votes for target areas that are likely to contain objects. Target areas are determined by placing  a log-polar grid, which we call the ``vote field,’’ centered at the voter location. The purpose of using a log-polar vote field is to reduce the spatial precision of the vote as the distance between voter location and target area increases. This is inspired by foveated vision systems found in nature, where the spatial resolution rapidly decreases from the fovea towards the periphery~\cite{land2009looking}. Once all visual evidence is  processed through voting, the accumulated votes are recorded in object presence maps, where the peaks indicate the presence of object instances.

Current state-of-the-art object detectors rely on local (or short-range) visual evidence to decide whether there is an object at that location (as in top-down methods) or an important keypoint such as a corner (as in bottom-up methods). On the other hand, HoughNet is able to integrate both short and long-range visual evidence through voting. An example is illustrated in Fig.~\ref{fig:teaser}, where the detected mouse gets strong votes from two keyboards, one of which is literally at the other side of the image. In another example (Fig.~\ref{fig:vis-results}, row 2, col 1), a ball on the right-edge of the image is voting for the baseball bat on the left-edge.  On the COCO  dataset, HoughNet achieves comparable results with the state-of-the-art bottom-up detector CenterNet~\cite{centernet2}, while being the fastest object detector among bottom-up detectors. It outperforms prominent one-stage (RetinaNet~\cite{retinanet}) and two-stage detectors (Faster RCNN~\cite{faster}, Mask RCNN~\cite{mask}).  To further show the effectiveness of our approach, we used the voting module of HoughNet in another task, namely, ``labels to photo’’ image generation. Specifically, we integrated the voting module to two different GAN models (CycleGAN~\cite{cyclegan} and Pix2Pix~\cite{pix2pix}) and showed that the performance is improved in both cases.

Our main contribution in this work is HoughNet, a voting-based bottom-up object detection method that is able to integrate near and long-range evidence for object detection. As a minor contribution, we created a mini training set called ``COCO \texttt{minitrain}’’, a curated subset of COCO \texttt{train2017} set, to reduce the computational cost of ablation experiments. We validated COCO \texttt{minitrain} in two ways by (i) showing that the COCO \texttt{val2017} performance of a model trained on COCO \texttt{minitrain} is strongly positively correlated with the performance of the same model trained on COCO \texttt{train2017}, and (ii) showing that COCO \texttt{minitrain} set preserves the object instance statististics. 
%We will publicly release HoughNet’s code and the \texttt{minitrain} dataset upon acceptance of this paper. 

\section{Related Work}

\paragraph{\textbf{Methods using log-polar fields/representations.}} Many biological systems have foveated vision  where the spatial resolution decreases from the fovea (point of fixation) towards the periphery. Inspired by this phenomenon, computer vision researchers have used log-polar fields for many different purposes including shape description \cite{shapematching}, feature extraction  \cite{akbas2017object} and foveated sampling/imaging \cite{roboticlogpolar}.

\paragraph{\textbf{Non-deep, voting-based object detection methods.}} In the pre-deep learning era, generalized Hough Transform (GHT) based voting methods have been used for object detection. The most influential  work  was  the Implicit Shape Model (ISM)~\cite{ism}. In ISM, Leibe \etal~\cite{ism} applied GHT for object detection/recognition and segmentation. During the training of the ISM, first, interest points are extracted and then a visual codebook (i.e. dictionary) is created using an unsupervised clustering algorithm applied on the patches extracted around interest points. Next, the algorithm matches the patches around each interest point to the visual word with the smallest distance. In the last step, the positions of the patches relative to the center of the object are associated with the corresponding visual words and stored in a table. During inference, patches extracted around interest points are matched to closest visual words. Each matched visual word casts votes for the object center. In the last stage, the location that has the most votes is identified, and object detection is performed using the patches that vote for this location. Later, ISM was further extended with discriminative frameworks~\cite{disc_hough, houghforest1, latent_hough, maxmargin, multiple_hough}. Okada~\cite{disc_hough} ensembled randomized trees using image patches as voting elements. Similarly, Gall and Lempitsky \cite{houghforest1} proposed to learn a mapping between image patches and votes using random forest framework. In order to fix the accumulation of inconsistent votes of ISM, Razavi~\etal~\cite{latent_hough} augmented the Hough space with latent variables to enforce consistency between votes.  In Max-margin Hough Transform~\cite{maxmargin}, Maji and Malik  showed the importance of learning visual words in a discriminative max-margin framework. Barinova~\etal~\cite{multiple_hough} detected multiple objects using energy optimization instead of non-maxima suppression peak  selection of ISM.

HoughNet is similar to ISM and its variants described above only at the idea level as all are voting based methods. There are two major differences: (i) HoughNet uses deep neural networks for part/feature (i.e. visual evidence) estimation, whereas ISM uses hand-crafted features; (ii) ISM uses a discrete set of visual words (obtained by unsupervised clustering) and each word’s vote is exactly known (stored in a table) after training. In HoughNet, however, there is not a discrete set of words and vote is carried through a log-polar vote field which takes into account the location precision as a function of target area.

\paragraph{\textbf{Bottom-up object detection methods.}}  Apart from the classical one-stage~\cite{ssd, yolo3, dssd, rrc, retinanet} vs. two-stage~\cite{faster, mask} categorization of object detectors, we can also categorize the current approaches into two: top-down and bottom-up. In the top-down approach~\cite{ssd, yolo3, retinanet, faster},  a near-exhaustive list of object hypotheses in the form of rectangular boxes are generated and objects are predicted in a  holistic manner based on these boxes. Designing the hypotheses space (e.g. parameters of anchor boxes) is a problem by itself~\cite{region_proposals}. Typically, a single template is responsible for the detection of the whole object. In this sense, recent anchor-free methods \cite{ fcos, centernet} are also top-down. On the other hand, in the bottom-up approach, objects \emph{emerge} from the detection of parts (or sub-object structures). For example, in CornerNet~\cite{cornernet}, top-left and bottom-right corners of objects are detected first, and then, they are paired to form whole objects. Following CornerNet, ExtremeNet~\cite{extremenet} groups extreme points  (e.g. left-most, etc.) and center points to form objects. Together with corner pairs of CornerNet~\cite{cornernet}, CenterNet~\cite{centernet2} adds center point to  model each object as a triplet. HoughNet follows the bottom-up approach based on a voting strategy: object presence score is voted (aggregated) from a wide area covering short and long-range evidence.

\paragraph{\textbf{Deep, voting-based object detection methods.}}   Qi \etal~\cite{3dhough} apply Hough voting for 3D object detection in point clouds. Sheshkus \etal~\cite{otherhoughnet}  utilize Hough transform for vanishing points detection in the documents.  For automatic pedestrian and car detection, Gabriel \etal~\cite{hough_proposal} proposed using discriminative generalized Hough transform for proposal generation in edge images, later to further refine the boxes, they fed these proposals to deep networks.  In the deep learning era, we are not the first to use a log-polar vote field in a voting-based model. Lifshitz~\etal~\cite{consensus} used a log-polar map to estimate keypoints for single person human pose estimation. Apart from the fact that~\cite{consensus} is tackling the human pose estimation task, there are several subtle differences. First, they prepare ground truth voting maps for each keypoint such that keypoints vote for every other one depending on its relative position in the log polar map. This requires manually creating static voting maps. Specifically, their model learns $H\times W\times R\times C$ voting map, where $R$ is the number of bins and $C$ is the augmented keypoints. In order to produce keypoint heatmaps they perform vote agregation at test phase. Second, this design restricts the model to learn only the keypoint locations as voters. When we consider the object detection task and its complexity, it is not trivial to decide the voters of the objects and prepare supervised static voting maps as in human pose estimation. Moreover, this design limits the voters to reside only inside of the object (e.g. person) unlike our approach where an object could get votes from far away regions. To overcome these issues, unlike their model we apply vote aggregation during training (they perform vote agregation only at test phase). This allows us to expose the latent patterns between objects and voters for each class. In this way, our voting module is able to get votes from non-labeled objects  (see the last row of Fig.~\ref{fig:vis-results}). To the best of our knowledge, we are the first to use a log-polar vote field in a voting-based deep learning  model to integrate the \textbf{long range interactions} for object detection.

Similar to HoughNet, Non-local neural networks (NLNN)~\cite{nlnn} and Relation networks (RN)~\cite{relation_networks} integrate long-range features.  As a fundamental difference, in NLNN, the relative displacement between interacting features is not taken into account. However, HoughNet uses this information encoded through the regions of the log-polar vote field. RN models object-object relations explicitly for proposal-based two-stage detectors.

\setlength\intextsep{8mm}
 \begin{figure}[t]
\includegraphics[width=\linewidth]{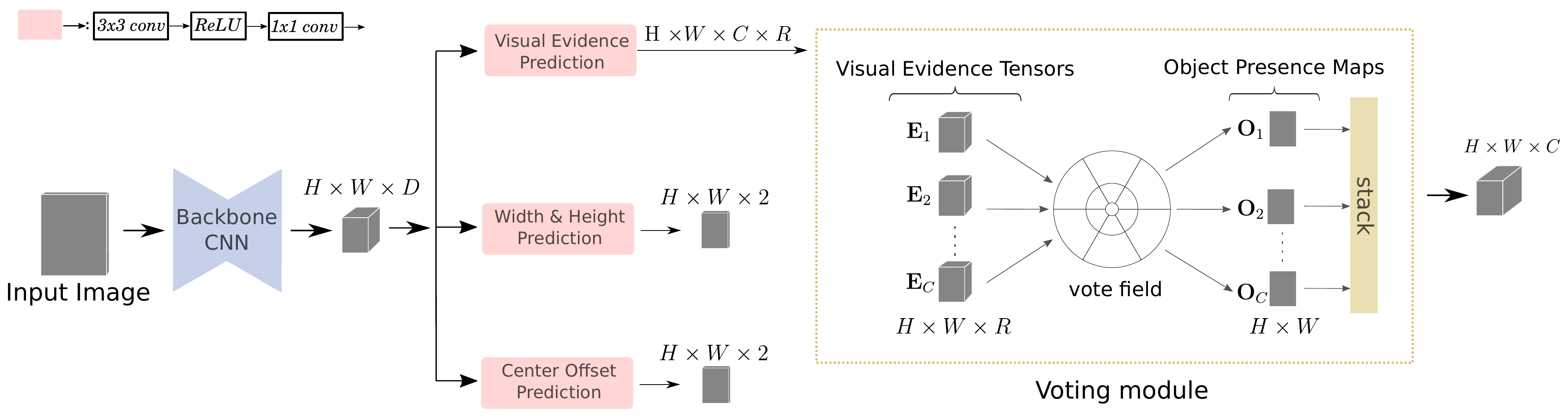}
\caption{Overall processing pipeline of HoughNet}
\label{fig:model}
\end{figure}
\section{HoughNet: the method and the models}

The overall processing pipeline of our method is illustrated in Fig.~\ref{fig:model}. To give a brief overview, the input image first passes through a backbone CNN, the output of which is connected to three different branches carrying out the predictions of (i) visual evidence scores, (ii) objects’ bounding box dimensions (width and height), and (iii) objects’ center location offsets. The first branch is where the voting occurs. Before we describe our voting mechanism in detail, we first introduce the log-polar vote field.

\subsection{The log-polar ``vote field’’}

\begin{wrapfigure}[22]{r}{0.4\textwidth}
\includegraphics[width=0.7\linewidth]{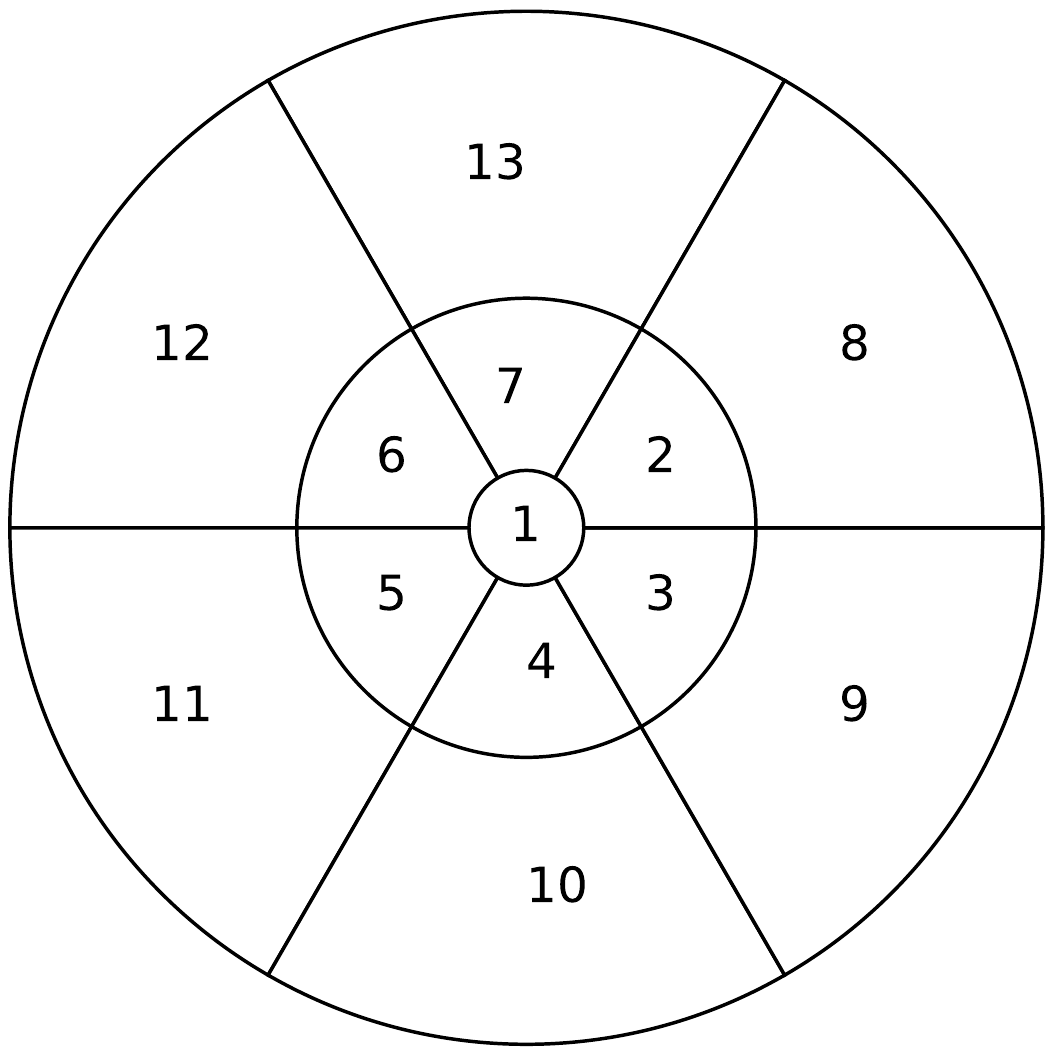}
\centering
\caption{A log-polar ``vote field’’ used in the voting module of HoughNet. Numbers indicate region ids. A vote field is parametrized by the number of angle bins, and the number and radii of eccentricity bins, or rings. In this particular vote field, there are a total of 13 regions, 6 angle bins and 3 rings. The radii of the rings are 2, 8 and 16, respectively
}
\label{fig:vf}
\end{wrapfigure}

We use the set of regions in a  standard log-polar coordinate system to define the regions through which votes are collected. A log-polar coordinate system is defined by the number and radii of eccentricity bins (or rings) and the number of angle bins. We call the set of cells or regions formed in such a coordinate system as the ``vote field’’ (Fig.~\ref{fig:vf}). In our experiments, we used  different vote fields with different parameters (number of angle bins, etc.) as explained in the Experiments section. In the following, $R$ denotes the number of regions in the vote field and $K_r$ is the number of pixels in a particular region $r$. $\Delta_r(i)$ denotes the relative spatial coordinates of the $i^{th}$ pixel in the $r^{th}$ region, with respect to the center of the field. We implement the vote field as a fixed-weight (non-learnable) transposed-convolution filter as further explained below.

\subsection{Voting module}

\setlength\intextsep{0pt}

After the  input image is passed through the backbone network and the ``visual evidence’’ branch, the voting module of HoughNet receives $C$ tensors $\mathbf{E}_1, \mathbf{E}_2, \dots, \mathbf{E}_C$, each of size  $H\times W\times R$, where $C$ is the number of classes, $H$ and $W$ are spatial dimensions and $R$ is the number of regions in the vote field. Each of these tensors contains class-conditional (i.e. for a specific class) ``visual evidence’’ scores. The job of the voting module is to produce $C$ ``object presence’’ maps $\mathbf{O}_1, \mathbf{O}_2, \dots, \mathbf{O}_C$, each of size $H\times W$. Then, peaks in these maps will indicate the presence of object instances. The voting process, which converts the visual evidence tensors (e.g. $\mathbf{E}_c$) to object presence maps (e.g. $\mathbf{O}_c$), works as described below.

Suppose we wanted to process the visual evidence at the $i^{th}$ row, $j^{th}$ column and the $r^{th}$ channel of an evidence tensor $\mathbf{E}$. When we place our vote field on a 2D map, centered at location $(i,j)$, the region $r$ marks the target area to be voted on, whose coordinates can be computed by adding the coordinate offsets $\Delta_r(\cdot)$ to $(i,j)$. Then, we add the visual evidence score $\mathbf{E}(i,j,r)$ to the target area of the object presence map. Note that this operation can be efficiently implemented using the ``transposed convolution’’ (or ``deconvolution’’) operation. Visual evidence scores from locations other than $(i,j)$ are processed in the same way and the scores are accumulated in the object presence map. We formally define this procedure in Algorithm \ref{algo:voting}, which takes in a visual evidence tensor as input and produces an object presence map\footnote{We provide a step-by-step animation of the voting process at \url{https://shorturl.at/ilOP2}.}.

\setlength\intextsep{8mm}
%\setlength\intextsep{0pt}
%\begin{wrapfigure}{r}{0.5\textwidth}
%\begin{minipage}{0.5\textwidth}
\begin{algorithm}
\caption{Vote aggregation algorithm}
\label{algo:voting}
\begin{algorithmic}
\REQUIRE Visual evidence tensor $\mathbf{E}_c$, Vote field relative coordinates $\Delta$
\ENSURE Object presence map $\mathbf{O}_c$

\STATE Initialize $\mathbf{O}_c$ with all zeros
\FOR{each pixel $(i,j,r)$ in $\mathbf{E}_c$} 
\STATE /*  $K_r$: number of pixels in the vote field region $r$ */
\FOR{$k$ = 1 to $K_r$} 
\STATE $(y,x) \leftarrow (i,j) + \Delta_r(k)$ 
\STATE $\mathbf{O}_c(y,x) \leftarrow \mathbf{O}_c(y,x) + \frac{1}{K_r} \mathbf{E}_c(i,j,r)$
\ENDFOR
\ENDFOR
\end{algorithmic}
\end{algorithm}
%\end{minipage}
%\end{wrapfigure}

\subsection{Network architecture}
Our network architecture design follows that of ``Objects as Points’’ (OAP)~\cite{centernet}. HoughNet consists of a backbone and three subsequent branches which predict  (i) visual evidence scores, (ii) bounding box widths and heights, and (iii) center offsets. Our voting module is attached to the visual evidence branch (Fig.~\ref{fig:model}).

The output of our backbone network is a feature map of size $H\times W\times D$, which is a result of inputting an image of size $4H\times 4W\times 3$. The backbone’s output is fed to all three branches. 
Each branch has one convolutional layer with $3\times 3$  filters followed by a ReLU layer  and another convolutional layer with $1\times 1$ filters. The visual evidence branch outputs $H\times W\times C\times R$ sized output where $C$ and $R$ correspond to the number of classes and vote field regions, respectively.
The width/height prediction branch outputs $H\times W\times 2$ sized output which predicts heights and widths for each possible object center.
Finally, center offset branch predicts relative displacement of center locations across the spatial axes. 
 
\paragraph{\textbf{Objective functions}} For the optimization of the visual evidence branch, we use the modified focal loss~\cite{retinanet} introduced in CornerNet~\cite{cornernet} (also used in~\cite{extremenet, centernet}).  In order to recover the lost precision of the center points due to down-sampling operations through the network, center offset prediction branch outputs class-agnostic local offsets of object centers. We optimize this branch using the  $L_{1}$ loss as the other bottom-up detectors~\cite{cornernet, extremenet, centernet} do. Finally, our width \& height prediction branch outputs class-agnostic width and height values of objects. For the optimization of this branch, we use  $L_{1}$ loss by scaling the loss by $0.1$ as proposed in OAP~\cite{centernet}. The overall loss is the sum of the losses from all branches.

\section{Experiments}
\label{sec:exp}

This section presents the experiments we conducted to show the effectiveness of our proposed method. First, we studied how different parameters of the vote field affect the final object detection performance. Next, we present several performance comparisons between HoughNet and the current state-of-the-art methods, on the COCO dataset. After presenting sample visual results for qualitative inspection, we describe our experiments on the ``labels to photo’’ task. We used PyTorch~\cite{pytorch} to implement HoughNet.

\paragraph{\textbf{Training and inference details}} We ran our experiments on 4 V100 GPUs. For training, we used $512\times512$ images unless stated otherwise. The training setup is not uniform across different experiments, mainly due to different backbones. %We explain the details of training processes in the related sections. 

The inference pipeline is common for all HoughNet models.  We extract center locations by applying a $3\times 3$ max pooling operation on object presence heatmaps and pick the highest scoring 100 points as detections. Then, we adjust these points using the predicted center offset values. Final bounding boxes are generated using the predicted width \& height values on these detections.
For testing, we follow the other bottom-up methods~\cite{extremenet, centernet, cornernet} and use two modes: (i) single-scale, horizontal-flip testing (SS testing mode), and (ii) multi-scale, horizontal-flip testing (MS testing mode). In MS, we use the following scale values, $0.6,1.0, 1.2, 1.5, 1.8$. To merge augmented test results, we use Soft-NMS~\cite{softnms}, and keep the top 100 detections. All tests are performed on a single V100 GPU. 
\subsection{Mini COCO}

For faster analysis in our ablation experiments, we created ``COCO \texttt{minitrain}’’ as a statistically validated mini training set. It is a subset of the COCO \texttt{train2017} dataset,  containing $25K$ images (about 20\% of \texttt{train2017}) and around $184K$ objects across $80$ object categories. We randomly sampled these images from the full set while preserving the following three quantities as much as possible: (i) proportion of object instances from each class, (ii) overall ratios of small, medium and large objects, (iii) per class ratios of small, medium and large objects.

To validate COCO \texttt{minitrain}, we computed the correlation between the \texttt{val2017} performance of a model when it is trained on \texttt{minitrain} with the same of when it is trained on \texttt{train2017}. Over  six different object detectors (Faster R-CNN, Mask R-CNN, RetinaNet, CornerNet, ExtremeNet and HoughNet), the Pearson correlation coefficients turned out to be $0.74$  and $0.92$ for \textit{AP}  and \textit{AP$_{50}$},  respectively. These values indicate strong positive correlation. Further details on  \texttt{minitrain} can be found at \url{https://github.com/giddyyupp/coco-minitrain}.

\subsection{Ablation experiments}
%\subsubsection{Voting field ablations}
\label{sec:vot_abl}
Here we analyze the effects  of the number of angle and ring bins of the vote field on performance. Models are trained on COCO \texttt{minitrain} and evaluated on \texttt{val2017} set with SS testing mode. The backbone is Resnet-101~\cite{resnet}. In order to get higher resolution feature maps, we add three deconvolution layers on top of the default Resnet-101 network, similar to~\cite{xiao2018simple}. We add $3\times3$ convolution filters before each  $4\times4$ deconvolution layer, and put batchnorm and ReLU layers after convolution and deconvolution filters.  We trained the network with a batch size of 44 for 140 epochs with Adam optimizer~\cite{adam}. Initial learning rate $1.75 \times 10^{-4}$ was divided by 10 at epochs 90 and 120.

\paragraph{\textbf{Angle bins}} We started with a large, $65$ by $65$, vote field with 5 rings. We set the radius of these rings from the most inner one to the most outer one as $2$, $8$, $16$, $32$ and $64$ pixels, respectively.  
We experimented with $60^\circ$, $90^\circ$, $180^\circ$ and $360^\circ$ bins. We do not split the center ring (i.e. region with id 1 in Fig.~\ref{fig:vf}) into further regions.  Results are presented in Table~\ref{table:abl_table}a. For the $180^\circ$ experiment, we divide the vote field horizontally. $90^\circ$ yields the best performance considering both  \textit{AP} and  \textit{AP$_{50}$}. We used this setting in the rest of the experiments.

%%%%%%%%%%%%%% Ablation lar tek tablo
\setlength\intextsep{0pt}
\begin{wraptable}{r}{0.5\linewidth}
%\begin{table}[ht]
\caption{Ablation experiments for the vote field. (a) Effect of angle bins on performance. Vote field with 90$^\circ$ has the best performance (considering \textit{AP} and  \textit{AP$_{50}$}). 
(b) Effect of central and peripheral regions. Here, the angle bin is 90$^\circ$ and the ring count is four. Disabling any of center or periphery hurts performance, cf. (a). (c) Effect of number of rings. Angle is 90$^\circ$  and vote field size is updated according to the radius of the last ring. Using 3 rings yields the best result. It is also the fastest model}
\centering
\resizebox{0.5\textwidth}{!}{
\begin{tabular}{lcccccccc}
\toprule % <-- Toprule here
Model & \textit{AP}& \textit{AP$_{50}$} &  \textit{AP$_{75}$}&  \textit{AP$_S$} & \textit{AP$_M$} & \textit{AP$_L$}  & FPS \\
\midrule % <-- Midrule here
60$^\circ$         & \textbf{24.6} & 41.3         & 25.0         & \textbf{8.2} & 27.7 & 36.2 & 3.4         \\
90$^\circ$     & \textbf{24.6}& \textbf{41.5} & 25.0         & \textbf{8.2} & 27.7 & 36.2 & \textbf{3.5} \\
180$^\circ$        & 24.5           & 41.1          & 24.8         & 8.1                 & 27.7 & \textbf{36.3} & \textbf{3.5}  \\
360$^\circ$        & \textbf{24.6} & 41.1         & \textbf{25.1} & 8.0   & \textbf{27.8} & \textbf{36.3} & \textbf{3.5} \\
 \midrule % <-- Midrule here
\multicolumn{8}{c}{(a) Varying the number of  angle bins}\\
\bottomrule % <-- Bottomrule here
% \shortstack{Model} & & \textit{AP}& \textit{AP$_{50}$} &  \textit{AP$_{75}$} &  \textit{AP$_S$} & \textit{AP$_M$} & \textit{AP$_L$}  & FPS        \\
% \midrule % <-- Midrule here
Only Center         & 23.8 & 39.5 & 24.5 & \textbf{7.9} & 26.8 & 34.7 & \textbf{3.5} \\
No Center         & \textbf{24.4} & \textbf{40.9} & \textbf{24.9} & 7.4 & \textbf{27.6} & \textbf{37.1} & 3.3 \\
Only Context &  23.6  & 39.7 & 24.2 & 7.4 & 26.4 & 35.9 & 3.4 \\
 \midrule % <-- Midrule here
\multicolumn{8}{c}{(b) Effectiveness of votes from center or periphery}\\
\bottomrule % <-- Bottomrule here
%\shortstack{Model} & \shortstack{Size} & \textit{AP}& \textit{AP$_{50}$} &  \textit{AP$_{75}$} &  \textit{AP$_S$} & \textit{AP$_M$} & \textit{AP$_L$}  & FPS \\
% \midrule % <-- Midrule here
5 Rings        & 24.6 & \textbf{41.5} & 25.0 & 8.2 & 27.7 & 36.2 & 3.5 \\
4 Rings        & 24.5 & 41.1 & 25.3 & 8.2 & \textbf{27.8} & 36.1 &   7.8 \\
3 Rings        & \textbf{24.8} & 41.3 & \textbf{25.6} & \textbf{8.4} & 27.6 & \textbf{37.5} & \textbf{15.6}\\
 \midrule % <-- Midrule here
\multicolumn{8}{c}{(c) Varying ring counts}\\
\bottomrule % <-- Bottomrule here
\end{tabular}}
\label{table:abl_table}
\end{wraptable}

\paragraph{\textbf{Effects of center and periphery}} We conducted experiments to analyze the importance of votes coming from different rings of the vote field. Results are presented in Table~\ref{table:abl_table}b. In the \textit{Only Center} case, we only keep the center ring and disable the rest. In this way, we only aggregate votes from features of the object center directly. This case corresponds to a traditional object detection paradigm where only local (short-range) evidence is used.  This experiment shows that votes from outer rings help  improve performance. For the \textit{No Center} case,  we only disable the  center ring. We observe that there is only $0.2 $ decrease in \textit{AP}. This suggests that the evidence  for successful detection is embedded mostly around the object center not directly inside the object center. In order to observe the power of long-range votes, we conducted another experiment called ``Only Context,’’ where we disabled the two most inner rings and used  only the three outer rings for vote aggregation. This model  reduced \textit{AP} by $1.0$ point compared to the full model.

\paragraph{\textbf{Ring count}} To find out how far an object should get votes from, we discard outer ring layers one by one as presented in Table~\ref{table:abl_table}c. The models with $5$ rings, $4$ rings and $3$ rings have $17$, $13$ and $9$ voting regions and $65$, $33$ and $17$ vote field sizes, respectively. The model with $3$ rings yields the best performance on \textit{AP} metric and is the fastest one at the same time. On the other hand, the model with $5$ rings yields $0.2$ \textit{AP$_{50}$} improvement over the model with $3$ rings.

From all these ablation experiments, we decided to use the model with 5 rings and $90^\circ$ as our \textit{Base Model}. Considering both speed and accuracy, we decided to use the model with $3$ rings and $90^\circ$ as our \textit{Light Model}.

\subsubsection{Voting module vs. dilated convolution}
\label{sec:comp_abl}

Dilated convolution~\cite{dilated}, which can include long-range features, could be considered as an alternative to our voting module. To compare performance, we trained models on \texttt{train2017} and evaluated them on \texttt{val2017} using the SS testing mode.

\setlength\intextsep{0pt}
\begin{wraptable}{r}{0.5\linewidth}
\caption{Comparing our voting module to an equivalent (in terms of number of parameters and the spatial filter size) dilated convolution filter on COCO \texttt{val2017} set. Models are trained on COCO \texttt{train2017} and results are presented on SS testing mode}
\centering
 \resizebox{0.5\columnwidth}{!}{
  \begin{tabular}{lccccccc}
    \toprule % <-- Toprule here
    \textbf{Method} & \textit{AP}& \textit{AP$_{50}$} &  \textit{AP$_{75}$} & \textit{AP$_S$} & \textit{AP$_M$} & \textit{AP$_L$} \\
    \midrule % <-- Midrule here
     Baseline  & 36.2 & 54.8 & 38.7 & 16.3 & 41.6 & 52.3         \\
     + Dilated Conv. & 36.6 & 56.1 & 39.2 & 16.7 & 42.0 & 53.6         \\
     + Voting Module         & \textbf{37.3} & \textbf{56.6}        & \textbf{39.9}        & \textbf{16.8}        & 
\textbf{42.6}        & \textbf{55.2}        \\
   \bottomrule 
  \end{tabular}}
 \label{table:dilated_compare}
\end{wraptable}

\textit{Baseline}: We consider OAP with ResNet-101-DCN backbone as baseline. The last $1 \times 1$ convolution layer of center prediction branch in OAP, receives  $H \times W \times D$ tensor and outputs object center heatmaps with a tensor of size  $H \times W \times C$.

\textit{Baseline  +  Voting Module}: We first adapt the last layer of center prediction branch in baseline to output $H \times W \times C \times R$ tensor, then attach our voting module on top of the center prediction branch. Adding the voting module increases parameters of the layer by $R$ times. The log-polar vote field is $65 \times 65$, and has 5 rings ($90^\circ$). With 5 rings and $90^\circ$ we end up with $R=17$ regions.

\textit{Baseline + Dilated Convolution}: We use dilated convolution with kernel size $4 \times 4$ and dilation rate 22 for the last layer of the center prediction branch in baseline. Using  $4 \times 4$ kernel increases parameters  16 times which is approximately equal to $R$ in the \textit{Baseline  +  Voting Module}. Using dilation rate 22, the filter size becomes  $67 \times 67$ which is close to $65 \times 65$ log-polar vote field.

%To make a fair comparison with  \textit{Baseline}, we train the \textit{Baseline +  Voting Module} and the \textit{Baseline + Dilated Convolution} utilizing deformable convolution filters before each deconvolution filter of ResNet-101 (i.e. Resnet-101-DCN).
%We trained the models on \texttt{train2017} and evaluated them  on  the \texttt{val2017} set. As shown in Table~\ref{table:dilated_compare}, the model with our voting module outperforms dilated convolution  for all \textit{AP} metrics.

For a fair comparison with  \textit{Baseline}, both \textit{Baseline +  Voting Module} and the \textit{Baseline + Dilated Convolution}
use Resnet-101-DCN backbone.
Our voting module outperforms dilated convolution  in all cases (Table~\ref{table:dilated_compare}).

\begin{table}
  \caption{HoughNet results on COCO \texttt{val2017} set for different training setups. $^\dagger$ indicates initialization with CornerNet weights, $^*$~indicates initialization with ExtremeNet weights. Results are given for SS and MS testing modes, respectively}
\centering
  \resizebox{1.0\textwidth}{!}{
  \begin{tabular}{lcccccccc}
    \toprule % <-- Toprule here
    Models & Backbone & \textit{AP} & \textit{AP$_{50}$} &  \textit{AP$_{75}$} & \textit{AP$_S$} & \textit{AP$_M$} & \textit{AP$_L$} & FPS\\
    \midrule % <-- Midrule here
Base & R-101 & 36.0  \textbf{/} 40.7 & 55.2   \textbf{/} 60.6 & 38.4  \textbf{/} 43.9 & 16.2  \textbf{/} 22.5 & 41.7 \textbf{/} 44.2 & 52.0  \textbf{/} 55.7  & 3.5  \textbf{/} 0.5         \\

Base & R-101-DCN & 37.3 \textbf{/} 41.6 & 56.6   \textbf{/} 61.2 & 39.9   \textbf{/} 44.9 & 16.8   \textbf{/} 22.6 & 42.6  \textbf{/} 44.8 & 55.2   \textbf{/} 58.8 & 3.3 \textbf{/} 0.4         \\

Light & R-101-DCN & 37.2   \textbf{/} 41.5 & 56.5  \textbf{/} 61.5 & 39.6   \textbf{/} 44.5 & 16.8 \textbf{/} 22.5 & 42.5   \textbf{/} 44.8        & 54.9  \textbf{/} 58.4         &  \textbf{14.3}   \textbf{/} 2.1           \\
Light & HG-104
&    40.9  \textbf{/}   43.7 &  59.2  \textbf{/}   61.9        &  44.1   \textbf{/} 47.3 &  23.8 \textbf{/}  27.5 &  45.3   \textbf{/}  45.9 & 52.6   \textbf{/}  56.2 &  6.1   \textbf{/}  0.8   \\

Light & HG-104$^*$
 &    41.7  \textbf{/}  44.7 &  60.5  \textbf{/}  63.2        &  45.6   \textbf{/} 48.9 &  23.9 \textbf{/}  28.0 &  45.7   \textbf{/}  47.0 & 54.6   \textbf{/}  58.1 &  5.9   \textbf{/}  0.8    \\
Light & HG-104$^*$
&   43.0   \textbf{/}   \textbf{46.1} &  62.2  \textbf{/}   \textbf{64.6}        &  46.9  \textbf{/}   \textbf{50.3} &  25.5 \textbf{/}   \textbf{30.0} &  47.6   \textbf{/}   \textbf{48.8} & 55.8   \textbf{/}   \textbf{59.7} &  5.7   \textbf{/}  0.8    \\

   \bottomrule % <-- Bottomrule here
  \end{tabular}}
 \label{table:coco-compare}
\end{table}

\subsection{Performance of HoughNet and comparison with baseline}
In Table~\ref{table:coco-compare}, we present the performance of HoughNet for different backbone networks, initializations and  our base-vs-light model, on the \texttt{val2017} set. There is a significant speed difference between Base and Light models. Our light model with R-101-DCN backbone is the fastest one (14.3 FPS) achieving $37.2$ \textit{AP} and $56.5$ \textit{AP$_{50}$}. We observe that initializing the backbone with a pretrained model improves the detection performance.  In Table~\ref{table:baseline-comp}, we compare HoughNet’s performance with its baseline OAP~\cite{centernet} for two different backbones. HoughNet is especially effective for small objects, it improves the baseline by 2.1 and 2.2 AP points for R-101-DCN and HG-104 backbones, respectively. We also provide results for the recently introduced \textit{moLRP}~\cite{lrp} metric, which combines localization, precision and recall in a single metric. Lower values are better.

%\setlength\intextsep{0pt}
%\begin{wraptable}{r}{0.5\linewidth}
\begin{table}
\caption{Comparison with baseline (OAP) on \texttt{val2017}. Results are given for single scale and multi scale test modes, respectively}
\centering
\resizebox{1.0\columnwidth}{!}{
 \begin{tabular}{ccccccccc}
   \toprule % <-- Toprule here
  {\textbf{Method} } &  \textit{AP}& \textit{AP$_{50}$} &  \textit{AP$_{75}$} & \textit{AP$_S$} & \textit{AP$_M$} & \textit{AP$_L$} & \textit{moLRP} $\downarrow$ \\
   \midrule % <-- Midrule here
   Baseline \textit{w} R-101-DCN  & 36.2  \textbf{/}  39.2 & 54.8  \textbf{/}  58.6  & 38.7   \textbf{/}  41.9 & 16.3  \textbf{/}  20.5& 41.6  \textbf{/}  42.6  & 52.3   \textbf{/}  56.2  &    71.1 \textbf{/}    68.3   \\
   + Voting Module     &  \textbf{37.2} \textbf{/} \textbf{41.5} & \textbf{56.5}   \textbf{/}\textbf{ 61.5} & \textbf{39.6}   \textbf{/} \textbf{44.5} & \textbf{16.8}   \textbf{/} \textbf{22.5} & \textbf{42.5}  \textbf{/} \textbf{44.8} & \textbf{54.9}   \textbf{/} \textbf{58.4}    &    \textbf{69.9} \textbf{/}    \textbf{66.6}  \\
   \midrule % <-- Midrule here
        Baseline  \textit{w} HG-104  & 42.2 \textbf{/}  45.1  & 61.1   \textbf{/}  63.5 & 46.0  \textbf{/}  49.3  & 25.2  \textbf{/} 27.8 & 46.4   \textbf{/}  47.7 & 55.2   \textbf{/}  \textbf{60.3}  &    66.1 \textbf{/}    63.9    \\
   + Voting Module     & \textbf{43.0}   \textbf{/}   \textbf{46.1 } &  \textbf{62.2}  \textbf{/}  \textbf{64.6} & \textbf{46.9}  \textbf{/}  \textbf{50.3} &  \textbf{25.5} \textbf{/}   \textbf{30.0} &  \textbf{47.6}   \textbf{/}   \textbf{48.8} & \textbf{55.8}   \textbf{/}   59.7  &     \textbf{65.6} \textbf{/}     \textbf{63.1}     \\
 
  \bottomrule
 \end{tabular}}
\label{table:baseline-comp}
\end{table}
%\end{wraptable}

%%%%%%%%%%%%%%%%% COCO Results:

%\setlength\intextsep{8mm}
%\setlength{\tabcolsep}{4pt}

\begin{table}[H]
\caption{Comparison with the state-of-the-art on COCO \texttt{test-dev}. The methods  are divided into three groups: two-stage, one-stage top-down and one-st
age bottom-up. The best results are boldfaced separately for each group.  Backbone names are shortened: R is ResNet, X is ResNeXt, F is FPN and HG is HourGlass. $^*$ indicates that the FPS values were obtained on the same AWS machine with a V100 GPU using the official repos in SS setup. The rest of the FPS are from their corresponding papers. F. R-CNN is Faster R-CNN
}
%Detection performances on COCO \texttt{test-dev} set. The methods  are divided into three groups: two-stage, one-stage top-down and one-stage bottom-up. The best results are boldfaced separately for each group.  We shorten the backbone names with the following mappings; R is ResNet, X is ResNeXt, F is FPN and HG is HourGlass. $^*$ in the FPS column indicates the FPS values that we obtained on the same AWS machine with a V100 GPU using the official repos in SS setup. The rest of the FPS results are from their corresponding papers. F. R-CNN is Faster R-CNN
\centering
 \resizebox{1.0\columnwidth}{!}{
 \begin{tabular}{llcccccccccc}
   \toprule % <-- Toprule here
  Method & Backbone & Initialize & Train size & Test size &  \textit{AP}& \textit{AP$_{50}$} &  \textit{AP$_{75}$} & \textit{AP$_S$} & \textit{AP$_M$} & \textit{AP$_L$} & FPS\\
   \midrule % <-- Midrule here
   \textit{Two-stage detectors:} & & & & & & & & & & & \\
%ION~\cite{ion} & VGG-16 & 23.6 & 43.2 & 23.6 &  6.4 & 24.1 & 38.3 \\
R-FCN~\cite{rfcn} & R-101 & ImageNet & 800$\times$800 & 600$\times$600 & 29.9 & 51.9 &  -  & 10.8 & 32.8 & 45.0 & 5.9 \\
CoupleNet~\cite{couplenet} & R-101 & ImageNet & ori. & ori. &  34.4 & 54.8 & 37.2 & 13.4 &  38.1 & 50.8 & -\\
F. R-CNN+++~\cite{resnet} & R-101 &  ImageNet &1000$\times$600 & 1000$\times$600& 34.9 & 55.7& 37.4& 15.6& 38.7 & 50.9 & - \\
F. R-CNN~\cite{fpn} & R-101-F & ImageNet &1000$\times$600 & 1000$\times$600& 36.2 & 59.1 & 39.0& 18.2 &39.0&48.2 & 5.0 \\
Mask R-CNN~\cite{mask} & X-101-F &  ImageNet & 1300$\times$800 & 1300$\times$800& 39.8 &  62.3 & 43.4 & 22.1 & 43.2 & 51.2 & 11.0 \\
Cascade R-CNN~\cite{cascade} & R-101 & ImageNet &  - & - & 42.8 & 62.1& 46.3 & 23.7& 45.5 &55.2& \textbf{12.0} \\
PANet~\cite{panet} & X-101 & ImageNet & 1400$\times$840 & 1400$\times$840& \textbf{47.4} & \textbf{67.2} & \textbf{51.8} & \textbf{30.1} & \textbf{51.7} & \textbf{60.0} & - \\
   \midrule % <-- Midrule here
\textit{One-stage detectors:} & & & & & & &  && &&\\
\textit{Top Down:} & & & & & & & && &&\\

SSD~\cite{ssd}         & VGG-16                 &  ImageNet & 512$\times$512 & 512$\times$512 & 28.8 & 48.5 & 30.3 & 10.9 & 31.8 & 43.5         & -\\
YOLOv3~\cite{yolo3}        & Darknet                  &  ImageNet & 608$\times$608 & 608$\times$608& 33.0 & 57.9 &  34.4 & 18.3 &  35.4 & 41.9         & \textbf{20.0}\\
DSSD513~\cite{dssd} & R-101         &  ImageNet & 513$\times$513 & 513$\times$513& 33.2& 53.3 & 35.2 & 13.0 & 35.4 &51.1        & -\\ 
RefineDet (SS)~\cite{refinedet} & R-101 & ImageNet & 512$\times$512 & 512$\times$512&  36.4 & 57.5 & 39.5 & 16.6 & 39.9 & 51.4 & - \\
RetinaNet~\cite{retinanet} & X-101-F & ImageNet & 1300$\times$800 & 1300$\times$800 & 40.8 & 61.1 & 44.1 & 24.1 & 44.2 & 51.2 & 5.4\\
RefineDet (MS)~\cite{refinedet} & R-101 & ImageNet & 512$\times$512 & $\leq$2.25$\times$ & 41.8 & 62.9 & 45.7 &  25.6 & 45.1 & 54.1 & - \\
OAP (SS)~\cite{centernet} & HG-104 & ExtremeNet &  512$\times$512 & ori. & 42.1 & 61.1 & 45.9 &  24.1 & 45.5 & 52.8 & 9.6$^*$ \\
FSAF (SS)~\cite{fsaf} & X-101 & ImageNet & 1300$\times$800 & 1300$\times$800 &   42.9  &  63.8 &  46.3 &  26.6 &  46.2  &  52.7 &  2.7 \\
FSAF (MS)~\cite{fsaf} & X-101 & ImageNet &  1300$\times$800 & $\sim\leq$2.0$\times$ &   44.6 &  65.2 &  48.6 &  29.7 &  47.1  &  54.6 &  - \\
FCOS~\cite{fcos}  & X-101-F  & ImageNet &  1300$\times$800 & 1300$\times$800 & 44.7 & 64.1 &  48.4 &  27.6 &  47.5 &    55.6 & 7.0$^*$ \\
FreeAnchor (SS)~\cite{freeanchor} & X-101-F & ImageNet &  1300$\times$960 & 1300$\times$960 & 44.9 & 64.3 & 48.5 & 26.8 & 48.3 & 55.9 & - \\
OAP (MS)~\cite{centernet} & HG-104 & ExtremeNet &  512$\times$512 & $\leq$1.5$\times$ & 45.1 & 63.9 & 49.3 &  26.6 & 47.1 & 57.7  & - \\
FreeAnchor (MS)~\cite{freeanchor} & X-101-F & ImageNet &   1300$\times$960 & $\sim\leq$2.0$\times$& \textbf{47.3} &   \textbf{66.3} &  \textbf{51.5} &   \textbf{30.6} & \textbf{50.4} &  \textbf{59.0} & - \\

\textit{Bottom Up:} & & & & & & & &&&& \\

ExtremeNet (SS)~\cite{extremenet} & HG-104 & - &  511$\times$511 & ori. & 40.2 & 55.5 & 43.2 & 20.4 & 43.2 & 53.1  & 3.0$^*$ \\
CornerNet (SS)~\cite{cornernet} & HG-104 & -  & 511$\times$511 & ori.& 40.5 & 56.5 & 43.1 & 19.4 & 42.7 & 53.9 & 5.2$^*$ \\
CornerNet  (MS)~\cite{cornernet}& HG-104 & -  & 511$\times$511 & $\leq$1.5$\times$ & 42.1 & 57.8 & 45.3 & 20.8 & 44.8 & 56.7 & - \\
ExtremeNet (MS)~\cite{extremenet} & HG-104 & - &  511$\times$511 & $\leq$1.5$\times$ & 43.7 & 60.5 & 47.0 & 24.1 & 46.9 & 57.6  & - \\
CenterNet (SS)~\cite{centernet2} & HG-104 & - &  511$\times$511 & ori. & 44.9 & 62.4 & 48.1 & 25.6 &  47.4 & 57.4 & 4.8$^*$ \\
CenterNet (MS)~\cite{centernet2} & HG-104 & - & 511$\times$511 & $\leq$1.8$\times$ &  \textbf{47.0} &  64.5 &  \textbf{50.7} & 28.9 &   \textbf{49.9} &  \textbf{58.9}  & - \\
\midrule % <-- Midrule here
HoughNet (SS)& HG-104 & -  & 512$\times$512 & ori. &  40.8 &   59.1 &   44.2 & 22.9 &  44.4 & 51.1 & \textbf{6.4}$^*$ \\
HoughNet (MS)& HG-104 & -  &  512$\times$512 & $\leq$1.8$\times$ & 44.0 &   62.4 &   47.7 & 26.4 &  45.4 & 55.2 & - \\
HoughNet (SS)& HG-104 &  ExtremeNet & 512$\times$512 & ori. &  43.1 & 62.2 & 46.8 & 24.6 &  47.0 & 54.4 & \textbf{6.4}$^*$  \\
HoughNet (MS)& HG-104 &  ExtremeNet &  512$\times$512 & $\leq$1.8$\times$ & 46.4 &  \textbf{65.1} &  \textbf{50.7} &  \textbf{29.1} &  48.5 &58.1 & -  \\
  \bottomrule % <-- Bottomrule here
\end{tabular}}
\label{table:stateoftheart-comp}
\end{table}

\subsection{Comparison with the state-of-the-art}
\label{sec:sota}

For comparison with the state-of-the-art, we use Hourglass-104~\cite{cornernet} backbone. We train Hourglass model with a batch size of 36 for 100 epochs using the Adam optimizer~\cite{adam}. We set the initial learning rate to  $2.5 \times 10^{-4}$ and divided  it by  10 at epoch 90.  Table~\ref{table:stateoftheart-comp} presents performances of HoughNet and several established state-of-the-art detectors. First,  we compare HoughNet with OAP~\cite{centernet} since it is the model on which we built HoughNet. In OAP, they did not present any results for “from scratch” training.  Instead they fine-tuned their model from ExtremeNet weights. When we do the same (i.e. initialize HoughNet with ExtremeNet weights), we obtain better results than OAP. However as expected, HoughNet is slower than OAP. Among the one-stage bottom-up object detectors, HoughNet performs on-par with the best bottom-up object detector by achieving $46.4$ \textit{AP} against $47.0$ \textit{AP} of CenterNet~\cite{centernet2}. HoughNet outperforms CenterNet on \textit{AP$_{50}$} ($65.1$ \textit{AP$_{50}$} vs. $64.5$ \textit{AP$_{50}$}). Note that, since our model is initialized with ExtremeNet weights, which makes use of the segmentation masks in its own training, our model  effectively uses more data compared to CenterNet. HoughNet is the fastest among one-stage bottom-up detectors. It is faster than CenterNet, CornerNet and more than twice as fast as ExtremeNet.

%In Table~\ref{table:baseline-comp}, we compare HoughNet’s performance with its baseline OAP~\cite{centernet}. HoughNet is especially effective for small objects, it improves the baseline by 2.1 and 2.2 AP points for R-101 \textit{w} DCN and HG-104 backbones, respectively.

%In Table~\ref{table:coco-compare}, we provide further results  for HoughNet, where we test different backbone networks, initializations and  our base-vs-light model. There is a significant speed difference between Base and Light models. Our light model with R-101 \textit{w} DCN backbone is the fastest one and runs at 14.3 FPS achieving $37.2$ \textit{AP} and $56.5$ \textit{AP$_{50}$}. We observe that initializing the backbone with a pretrained model improves the detection performance.  

%%%%%%%%%%% Figure GORSEL SONUCLAR %%%%%%%%
%\setlength\intextsep{8mm}
\begin{figure*}
\captionsetup[subfigure]{labelformat=empty}
\centering
\begin{subfigure}[b]{0.16\textwidth}
   \caption{Detection}
   \includegraphics[width=1.0\textwidth]{./figs/visual_res2/keyboard/15_det.jpg}
   \includegraphics[width=1.0\textwidth]{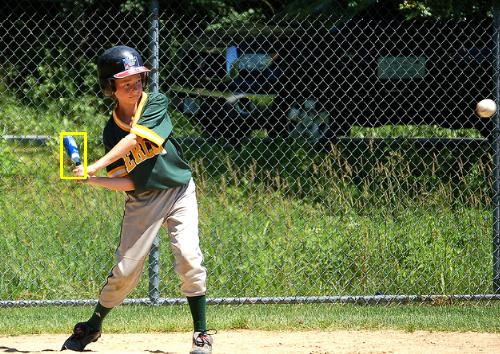}
   \includegraphics[width=1.0\textwidth]{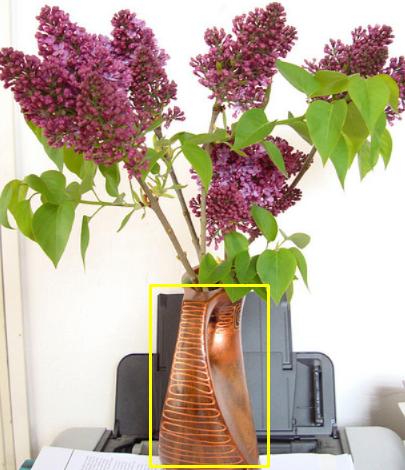}
   \includegraphics[width=1.0\textwidth]{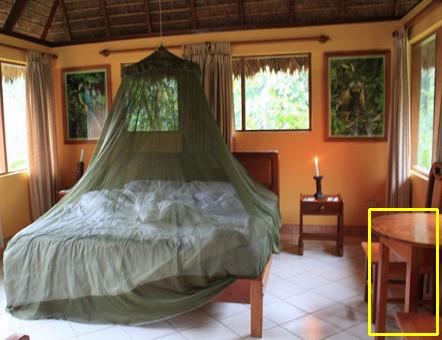}

\end{subfigure}
\begin{subfigure}[b]{0.16\textwidth}
   \caption{Voters}
   \includegraphics[width=1.0\textwidth]{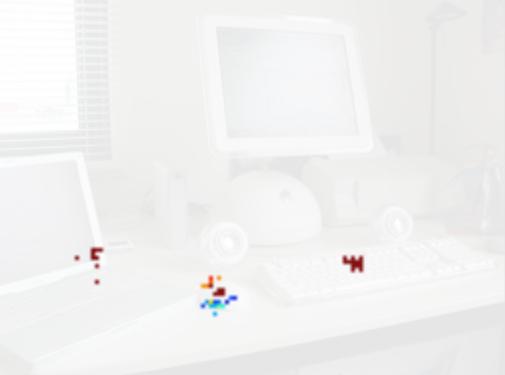}
\includegraphics[width=1.0\textwidth]{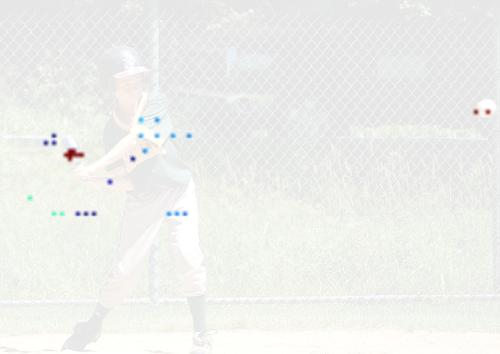}
  \includegraphics[width=1.0\textwidth]{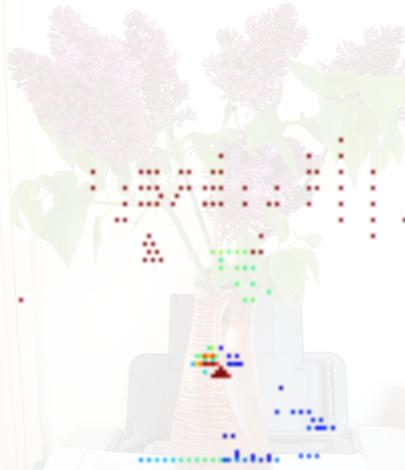}
   \includegraphics[width=1.0\textwidth]{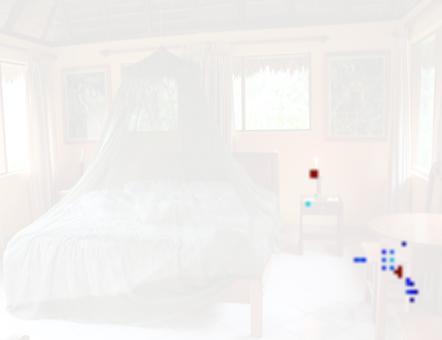}
\end{subfigure}
\begin{subfigure}[b]{0.16\textwidth}
   \caption{Detection}
   \includegraphics[width=1.0\textwidth]{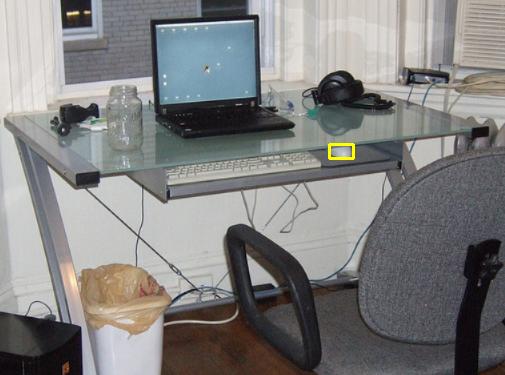}
   \includegraphics[width=1.0\textwidth]{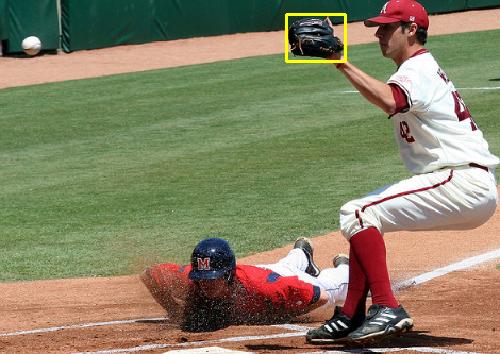}
   \includegraphics[width=1.0\textwidth]{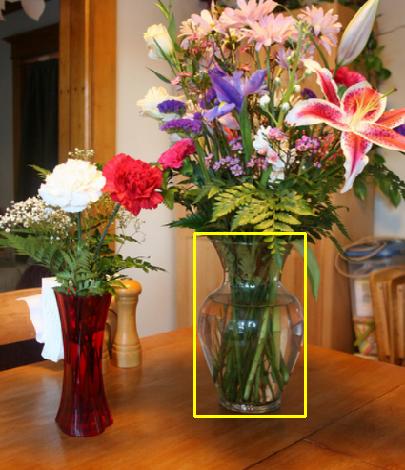}
   \includegraphics[width=1.0\textwidth]{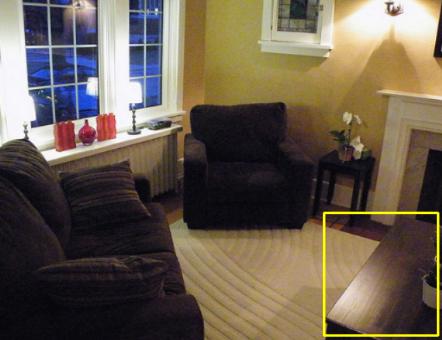}
\end{subfigure}
\begin{subfigure}[b]{0.16\textwidth}
   \caption{Voters}
   \includegraphics[width=1.0\textwidth]{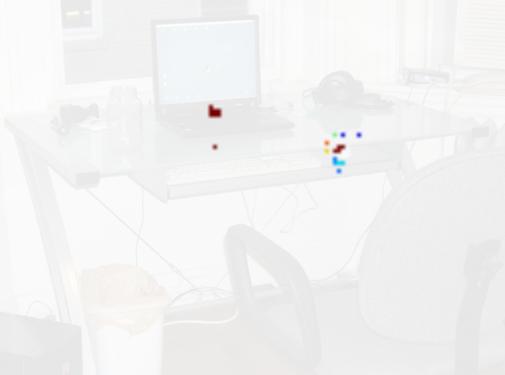}
\includegraphics[width=1.0\textwidth]{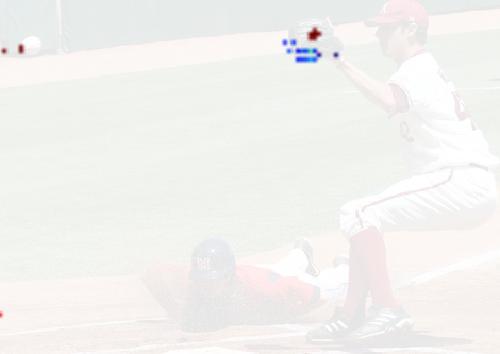}
  \includegraphics[width=1.0\textwidth]{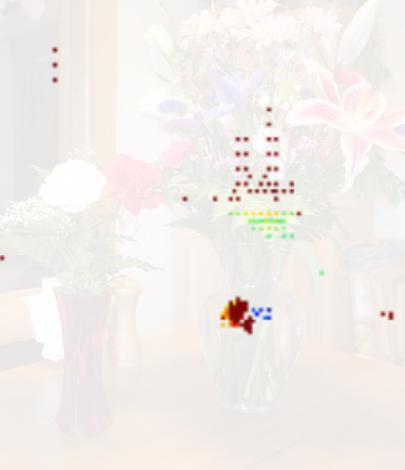}
  \includegraphics[width=1.0\textwidth]{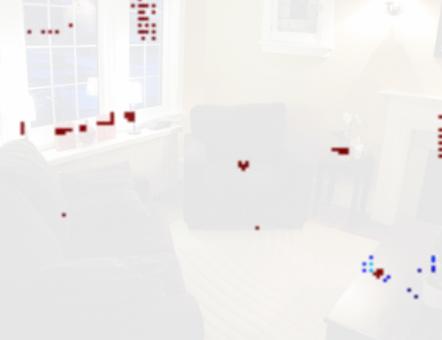}
\end{subfigure}
\begin{subfigure}[b]{0.16\textwidth}
   \caption{Detection}
   \includegraphics[width=1.0\textwidth]{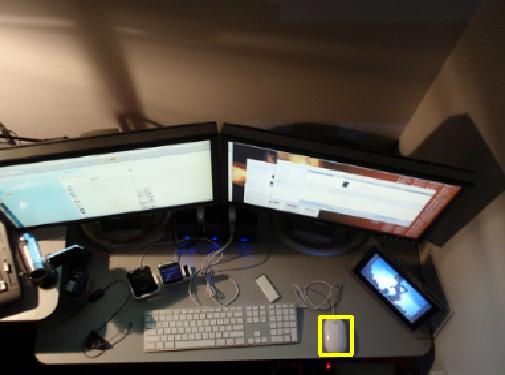}
   \includegraphics[width=1.0\textwidth]{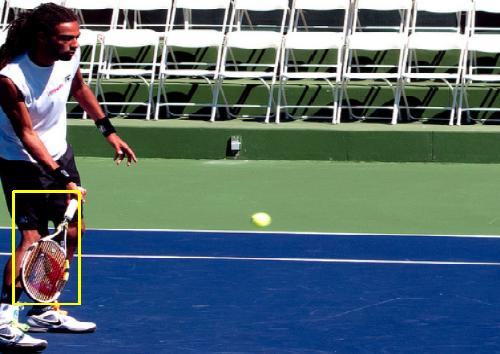}
   \includegraphics[width=1.0\textwidth]{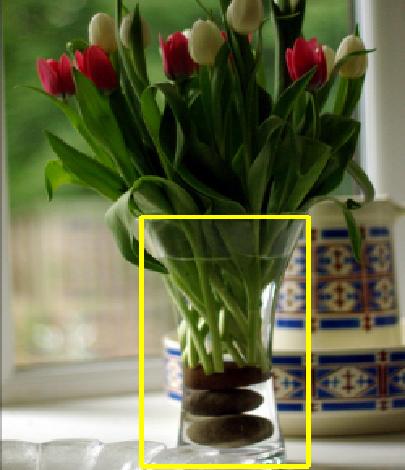}
  \includegraphics[width=1.0\textwidth]{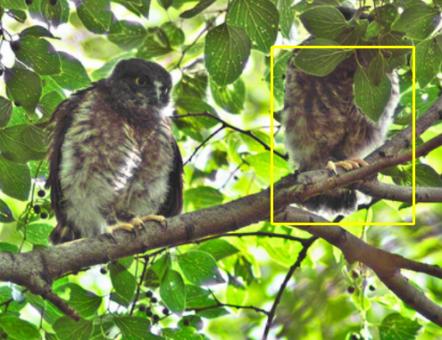}
\end{subfigure}
\begin{subfigure}[b]{0.16\textwidth}
   \caption{Voters}
   \includegraphics[width=1.0\textwidth]{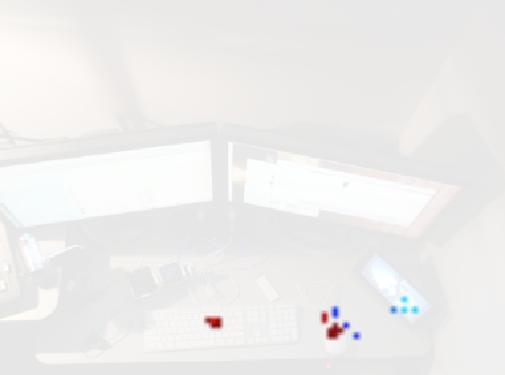}
\includegraphics[width=1.0\textwidth]{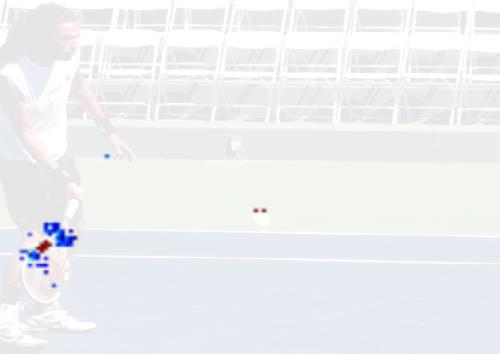}
  \includegraphics[width=1.0\textwidth]{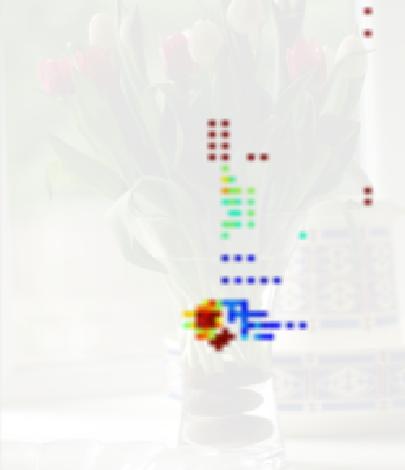}
  \includegraphics[width=1.0\textwidth]{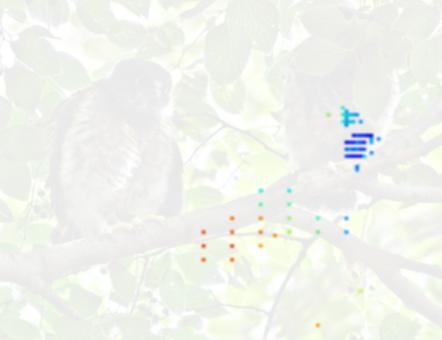}
\end{subfigure}
\caption{Sample detections of HoughNet and their vote maps. In the ``detection’’ columns, we show a correctly detected object, marked with a yellow bounding box. In the ``voters’’ columns, the locations that vote for the detection are shown. Colors indicate vote strength based on the standard ``jet’’ colormap (red is high, blue is low; Fig.~\ref{fig:teaser}).  In the \textbf{top row}, there are three ``mouse’’ detections. In all cases, in addition to the local votes (that are on the mouse itself), there are strong votes coming from nearby ``keyboard’’ objects. This voting pattern is justified given that mouse and keyboard objects frequently co-appear.  A similar behavior is observed in the detections of ``baseball bat’’, ``baseball glove’’ and ``tennis racket’’ in the \textbf{second row}, where they get strong votes from ``ball’’ objects that are far-away. Similarly, in the \textbf{third row}, ``vase’’ detections get strong votes from the flowers. In the first example of the \textbf{bottom row}, ``dining table’’ detection gets strong votes from the candle object, probably because they co-occur frequently. Candle is not among the 80 classes of COCO dataset. Similarly, in the second example in the  \textbf{bottom row}, ``dining table’’  has strong votes from objects and parts of a standard living room. In the last example, partially occluded bird gets strong votes (stronger than the local votes on the bird itself) from the tree branch}
\label{fig:vis-results}
\end{figure*}

We provide visualization of votes for sample detections of HoughNet for qualitative visual inspection (Fig.~\ref{fig:vis-results}). These detections clearly show that HoughNet is able to make use of long-range visual evidence.

\setlength\intextsep{0pt}   
\begin{wraptable}[9]{r}{0.50\textwidth}
\caption{Comparison of FCN scores for the ``labels to photo’’ task on the  Cityscapes~\cite{cityscapes} dataset}

\resizebox{0.5\columnwidth}{!}{
 \begin{tabular}{lccc}
   \toprule % <-- Toprule here
   \textbf{Method} & \textit{Per-pixel acc.}& \textit{ Per-class acc.} &  \textit{Class IOU } \\
   \midrule % <-- Midrule here
    CycleGAN  & 0.43   & 0.14 & 0.09           \\
    + Voting & \textbf{0.52} & \textbf{0.17} & \textbf{0.13} \\
    \midrule % <-- Midrule here
    pix2pix         & 0.71 &  \textbf{0.25} & 0.18 \\
    + Voting & \textbf{0.76}  &  \textbf{0.25}        & \textbf{0.20}  \\
  \bottomrule % <-- Bottomrule here
 \end{tabular}}
\label{table:gans}
\end{wraptable}

\subsection{Using our voting module in another task}
One task where long-range interactions could be useful is the task of image generation from a given label map. There are two main approaches to solve this task; using unpaired and paired data for training. We take CycleGAN~\cite{cyclegan} and Pix2Pix~\cite{pix2pix} as our baselines for unpaired and paired approaches, respectively. We attach our voting module at the end of CycleGAN~\cite{cyclegan} and Pix2Pix~\cite{pix2pix} models.

\begin{figure}
\captionsetup[subfigure]{labelformat=empty}
\centering
\hfill
\begin{subfigure}[b]{0.14\textwidth}
   \caption{Input}
   \includegraphics[width=1.0\textwidth]{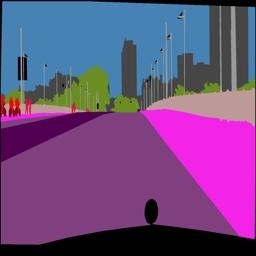}
   \includegraphics[width=1.0\textwidth]{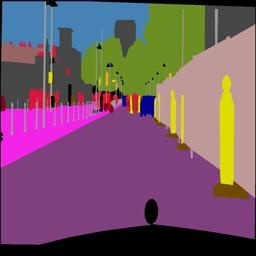}
   \includegraphics[width=1.0\textwidth]{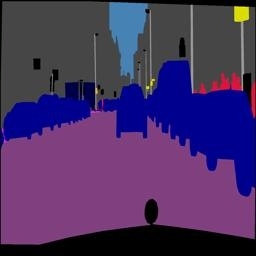}
\end{subfigure}
\begin{subfigure}[b]{0.14\textwidth}
   \caption{CycleGAN}
   \includegraphics[width=1.0\textwidth]{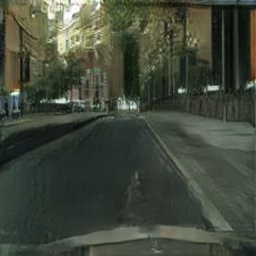}
   \includegraphics[width=1.0\textwidth]{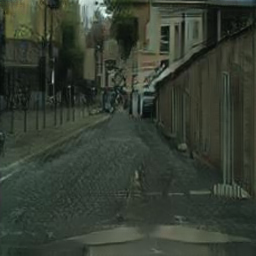}
   \includegraphics[width=1.0\textwidth]{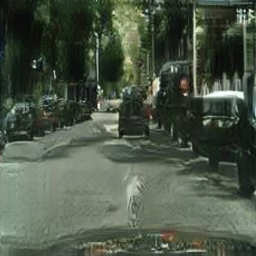}
\end{subfigure}
\begin{subfigure}[b]{0.14\textwidth}
   \caption{\shortstack{ + Voting}}
   \includegraphics[width=1.0\textwidth]{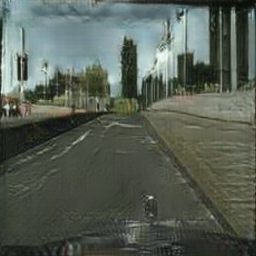}
   \includegraphics[width=1.0\textwidth]{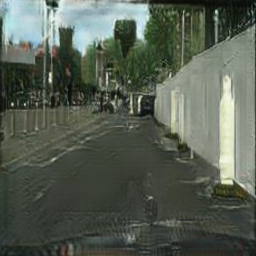}
   \includegraphics[width=1.0\textwidth]{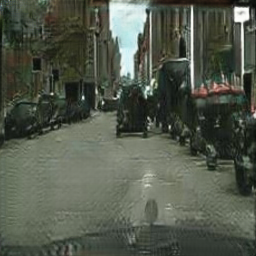}
\end{subfigure}
\hfill
\begin{subfigure}[b]{0.14\textwidth}
    \caption{Input}
   \includegraphics[width=1.0\textwidth]{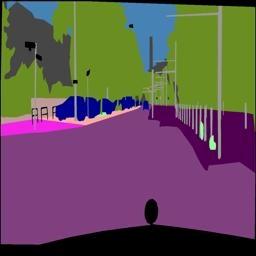}
   \includegraphics[width=1.0\textwidth]{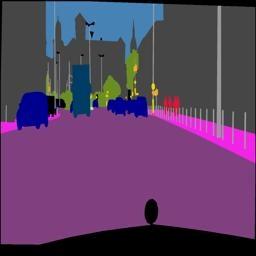}
   \includegraphics[width=1.0\textwidth]{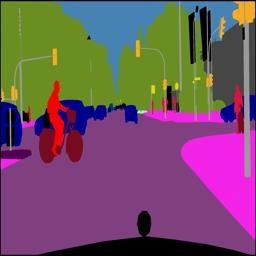}
\end{subfigure}
\begin{subfigure}[b]{0.14\textwidth}
   \caption{Pix2Pix}
   \includegraphics[width=1.0\textwidth]{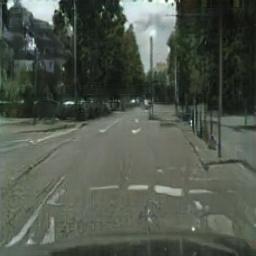}
   \includegraphics[width=1.0\textwidth]{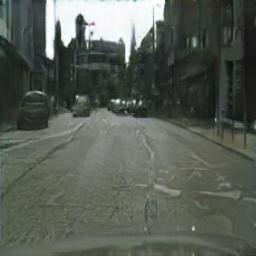}
   \includegraphics[width=1.0\textwidth]{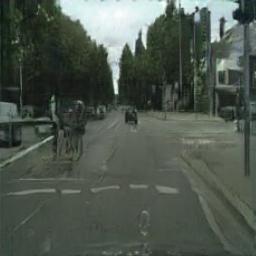}
\end{subfigure}
\begin{subfigure}[b]{0.14\textwidth}
   \caption{\shortstack{ + Voting}}
   \includegraphics[width=1.0\textwidth]{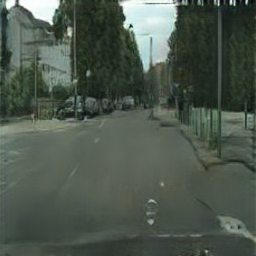}
   \includegraphics[width=1.0\textwidth]{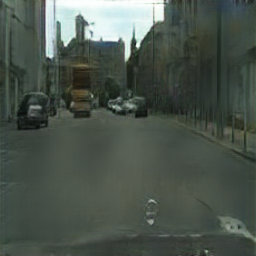}
   \includegraphics[width=1.0\textwidth]{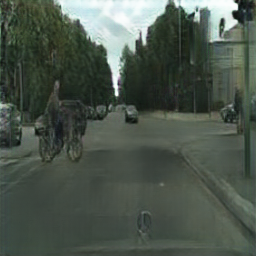}
\end{subfigure}
~~~~
\caption{Sample qualitative results for the ``labels to photo’' task. When integrated with  CycleGAN, our voting module helps generate better images in the sense that the image conforms to the input label map better. In all three images, CycleGAN fails to generate sky, buildings and falsely generates vegetation in the last image. When used with Pix2Pix, it helps generate more detailed images. In the first row, cars and buildings can be barely seen for Pix2Pix. Similarly, a bus is generated as a car and a bicycle is silhouetted in the second and third images, respectively. Our voting module fixes these errors}
\label{fig:vis-results-gan}
\end{figure}

For quantitative comparison, we use the Cityscapes~\cite{cityscapes} dataset. In Table~\ref{table:gans}, we present FCN scores~\cite{fcn} (which is used as the measure of success in this task) of CycleGAN and Pix2Pix with and without our voting module. To obtain the ``without’’ result, we used the already trained model shared by the authors. We obtained the ``with” result using the official training code from their repositories. In both cases evaluation was done using the official test and evaluation scripts from their repos.  Results show that using the voting module improves FCN scores by large margins.
Qualitative inspection also shows that when our voting module is attached, the generated images conform to the given input segmentation maps better (Fig.~\ref{fig:vis-results-gan}). This is the main reason for the quantitative improvement. Since Pix2Pix is trained with paired data, generated images follow input segmentation maps, however, Pix2Pix fails to generate small details.

%%%%%%%%%%% Figure GAN GORSEL SONUCLAR %%%%%%%%
\setlength\intextsep{8mm}
\section{Conclusion}
% discussion olarak yazilabilecek seyler: top-down vs bottom-up isimlendirmeleri. long-range interaction’larin nasil saglanabilecegi. context model’leri ile iliskiler. effective receptive field’in sota detector’lar icin olculmesi. 
% dicussion: voting module’u ekleyince learnable parametre sayisi artmiyor ama FLOP artiyor. 
% bottom-up’lara one-stage diyoruz ama two-stage de denebilir. cunku grouping var. 
% minicoco figure’unde ap ve ap50 siralamasinin farkli cikmasi 
% deconv filtresi learnable mi olmali? 

In this paper, we presented HoughNet, a new, one-stage, anchor-free, voting-based, bottom-up object detection method.  HoughNet determines the presence of an object at a specific location by the sum of the votes cast on that location. Voting module of HoughNet is able to use both short and long-range evidence through its log-polar vote field. Thanks to this ability, HoughNet generalizes and enhances current object detection methodology, which typically relies on only local (short-range) evidence.  We show that HoughNet performs on-par with the state-of-the-art  bottom-up object detector, and obtains comparable results with one-stage and two-stage methods. To further validate our proposal, we used the voting module of HoughNet in an image generation task. Specifically, we showed that our voting module significantly improves the performance of two GAN models in a ``labels to photo’’ task.

\paragraph*{\textbf{Acknowledgments}}
This work was supported by the Scientific and Technological Research Council of Turkey (T\"{U}B\.{I}TAK) through the project titled ``Object Detection in Videos with Deep Neural Networks" (grant \#117E054). The numerical calculations reported in this paper were partially performed at T\"{U}B\.{I}TAK ULAKB\.{I}M, High Performance and Grid Computing Center (TRUBA resources). We also gratefully acknowledge the support of the AWS Cloud Credits for Research program.

\setlength\intextsep{8mm}

\clearpage

{\small
\bibliographystyle{splncs04}
\bibliography{references}
}
\end{document}